\documentclass[runningheads]{llncs}
\usepackage{graphicx}
\usepackage{amsmath,amssymb} 
\usepackage{color}
\usepackage[width=122mm,left=12mm,paperwidth=146mm,height=193mm,top=12mm,paperheight=217mm]{geometry}
\usepackage{algorithm,algorithmic,booktabs}
\begin{document}
\pagestyle{headings}
\mainmatter
\def\ECCV18SubNumber{***}  

\title{Fast Dynamic Routing Based on Weighted Kernel Density Estimation} 

\titlerunning{ECCV-18 submission ID \ECCV18SubNumber}

\authorrunning{ECCV-18 submission ID \ECCV18SubNumber}

\author{Suofei Zhang$^{1}$, Wei Zhao$^{2}$, Xiaofu Wu$^{1}$, Quan Zhou$^{1}$}
\institute{$^1$Nanjing University of Post and Telecommunication\\$^{2}$SIAT, Chinese Academy of Sciences}

\maketitle

\begin{abstract}
Capsules as well as dynamic routing between them are most recently proposed structures for deep neural networks. A capsule groups data into vectors or matrices as poses rather than conventional scalars to represent specific properties of target instance. Besides of pose, a capsule should be attached with a probability (often denoted as activation) for its presence. The dynamic routing helps capsules achieve more generalization capacity with many fewer model parameters. However, the bottleneck that prevents widespread applications of capsule is the expense of computation during routing. To address this problem, we generalize existing routing methods within the framework of weighted kernel density estimation, and propose two fast routing methods with different optimization strategies. Our methods prompt the time efficiency of routing by nearly 40\% with negligible performance degradation. By stacking a hybrid of convolutional layers and capsule layers, we construct a network architecture to handle inputs at a resolution of $64\times{64}$ pixels. The proposed models achieve a parallel performance with other leading methods in multiple benchmarks.

\end{abstract}

\section{Introduction}

During the last decade, deep learning algorithms, especially Convolutional Neural Networks (CNNs) have achieved remarkable progress on numerous practical vision tasks~\cite{Krizhevsky2012ImageNet,karpathy2014large,lecun2015deep}. However, the stack of convolutional filters and non-linearity units still implies difficulty of understanding the internal organization of neural networks. It brings a reputation of ``black box'' to current neural networks~\cite{alain2016understanding}. Conversely, human vision systems always show much higher interpretability during the procedure of recognition. We can explicitly tell the specific cues such as shape, color, texture or intermediate semantic concepts, and build the part-whole relationship as proofs to our final decisions. The capsule structure introduces an analogous way to construct neural networks with higher interpretability. It uses grouped scalars as pose to represent specific properties of current part. Based on the multi-dimensional representation, it finds clusters within a ``routing-by-agreement'' framework as an interpretable representation during forward inference of network.
\begin{figure}
\centering
\includegraphics[width=12cm]{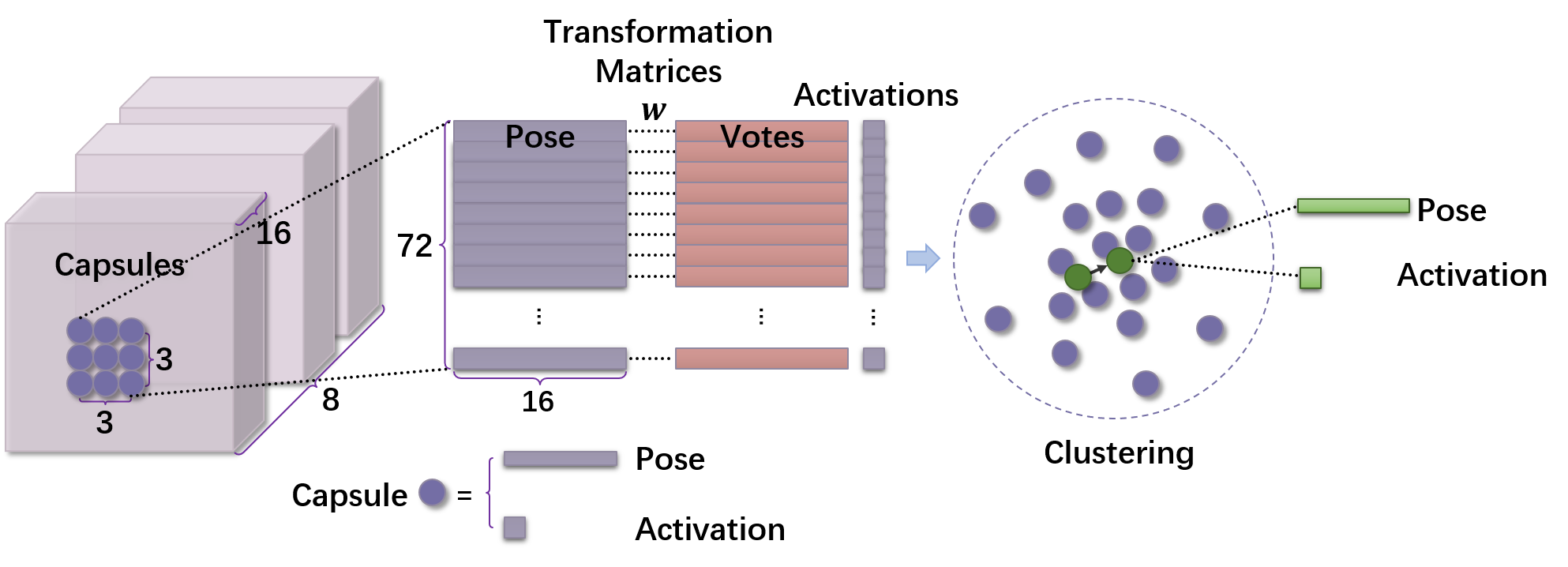}
\caption{Capsule structures and dynamic routing between them. A capsule consists of a group of 16 scalars as pose, as well as an 1D activation as magnitude. The pose of capsule can have the format as either vector~\cite{sabour2017dynamic} or matrix~\cite{hinton2018emrouting}. Here routing between capsules occurs in a [3,3] grid including a sum of 72 capsules. It finally results in new capsules with their activations from clusters of candidate samples as output of whole procedure. Here only one output capsule is illustrated.}
\label{fig:routing}
\end{figure}

Dynamic routing has been proven as an effective approach with higher generalization capacity and fewer parameters~\cite{hinton2018emrouting,zhaoweiemnlp2018}. However, it relies on intensive computation of clustering during inference. The expense of computation prevents capsules from widespread applications in practical vision tasks. To address this problem, we model dynamic routing from the perspective of nonparametric clustering and Kernel Density Estimation (KDE), proposing a target function to explain the procedure of routing. The main contributions of this paper are twofold: (1) We equip two simplified routing methods with different optimization strategies. Comparisons in different benchmarks empirically prove that our fast methods significantly prompt the efficiency of routing with negligible performance degradation. (2) We propose a hybrid network structure consisting of both convolutional layers and capsules. The network can efficiently handle images with a resolution of $64\times{64}$ pixels. Experiments in multiple benchmarks show that our models achieve a parallel performance with other leading methods. We believe our research of simplified routing methods prompts the potentiality of capsule networks to be extensively applied in practical vision tasks.

\subsection{Dynamic routing between capsules}
Capsule structure is originally proposed in~\cite{hinton2011transforming}. Differing from a convolutional neural unit, each layer here is divided into small groups of neurons called \textit{capsules}. As an instance in Figure~\ref{fig:routing}, an 128D vector of data is grouped into 8 capsules with 16D data as poses. Normally, an 1D activation is combined with the pose to represent the magnitude of current capsule. Based on such multi-dimensional representation, a dynamic routing mechanism over capsules~\cite{sabour2017dynamic} is designed between layers. 

Dynamic routing is a reminiscent of the unsupervised clustering, which is purely based on the divergence between samples. Hence, different clustering methods were considered here as nonparametric routing framework~\cite{hinton2018emrouting}. Finally, the activations of capsules can be exploited for solving typical machine learning problems directly. Meanwhile, the pose of capsule can also be mapped to some specific properties as regularization. The capsule structure comes with attractive features such as higher interpretability and generalization capacity. However, the high complexity of clustering method also leads to low efficiency for both training and inference. This drawback prevents the structure from applications on large scale vision tasks such as ImageNet~\cite{ILSVRC15}. 

\section{Kernel density estimation}

KDE, also known as Parzen Window method, is a nonparametric technique for describing underlying empirical distribution over given samples. Since the calculation of KDE is only related to specific form of kernel function and distance metric, it normally leads to less computation as well as higher efficiency. Given $n$ samples $\{\boldsymbol{u}_i|i=1,\dots,n\}$ in the $D$-dimensional space $\mathbb{R}^D$, the multivariate KDE of random variable $\boldsymbol{v}$ can be defined as
\begin{align}
\label{eq:kde}
\hat{f}(\boldsymbol{v}) \triangleq \frac{1}{nz_k}\sum\limits_{i=1}^nk(d(\boldsymbol{v}-\boldsymbol{u}_i)),
\end{align}
where $k(x)$ is a bounded function called \textit{profile} function with support in univariate space $\mathbb{R}$. $z_k$ is the normalization constant only related to specific $k(\cdot)$. Some instances of $k(x)$ can be illustrated as~\cite{wand1994kernel}
\begin{align}
Gaussian: k(x)\triangleq\exp{(-\frac{x}{2})}, && Epanechnikov: k(x)\triangleq\left\{\begin{aligned}1-x && x\in{[0,1)}\\ 0 && x\ge{1}.\end{aligned}\right.
\end{align}
The distance metric between samples $d(\boldsymbol{v}-\boldsymbol{u}_i)$ can be replaced by different definitions of distance, e.g., $\ell_2$ norm: $d(\boldsymbol{v}-\boldsymbol{u}_i)=||\boldsymbol{v}-\boldsymbol{u}_i||^2$, and Mahalanobis distance: $(\boldsymbol{v}-\boldsymbol{u}_i)^T\Sigma^{-1}(\boldsymbol{v}-\boldsymbol{u}_i)$. Intuitively, $\hat{f}(\boldsymbol{v})$ reflects the average distance between current point $\boldsymbol{v}$ and surrounding samples.

\section{Dynamic routing based on weighted KDE}

KDE is the basic technique of various clustering methods~\cite{schwander2013learning}. Despite the relationship between dynamic routing and clustering, there still exists two major problems preventing using KDE for dynamic routing directly. First, eq.~\ref{eq:kde} only considers the case of density estimation for one cluster, rather than routing between samples and multiple clusters. Second, there is no mechanism for including activation in the framework of KDE. To address these problems, we extend the density estimation from one cluster to mixture of clusters as
\begin{align}
\label{eq:kdem}
\hat{f}(\boldsymbol{v},\boldsymbol{r})\triangleq\frac{1}{n_lz_k}\sum\limits_{j=1}^{n_{l+1}}\sum\limits_{i=1}^{n_l}r_{ij}a_i^uk(d(\boldsymbol{v}_j-\boldsymbol{u}_i)),
\end{align}
where $n_l$ and $n_{l+1}$ are number of capsules at layer $l$ and $l+1$, respectively. We rewrite $\hat{f}(\boldsymbol{v})$ as $\hat{f}(\boldsymbol{v},\boldsymbol{r})$ given weights $\boldsymbol{r}$ as parameters of model. $\{\boldsymbol{v}_j|j=1,\dots,n_{l+1}\}$ are poses of capsule at layer $l+1$, i.e. the resulting clusters, while $\{\boldsymbol{u}_i|i=1,\dots,n_l\}$ are candidate samples at layer $l$. Note that as shown in Figure~\ref{fig:routing}, we actually use the transformed votes $\boldsymbol{u}_{i|j}$ instead of $\boldsymbol{u}_i$ for clustering. However, since the clustering only takes place in the scope of $\boldsymbol{v}_j$ and its corresponding votes $\{\boldsymbol{u}_{i|j}|\quad{i}|j=1,\dots,n_l\}$, such change of notation will not break the following derivation. We will still use $\boldsymbol{u}_i$ for simplicity. Activation $a_i^u$ of input capsule is introduced here as prior knowledge from below layer. It is a straightforward way to let samples with higher activations give more impact to final position of cluster. We introduce $r_{ij}$ here to measures how much $\boldsymbol{u}_i$ contributes to the position of $\boldsymbol{v}_j$, namely routing weight between 2 capsules. 

Since in the case of dynamic routing, clustering jointly takes place between samples and multiple clusters now. For the diversity of clusters, we want one sample can contribute to clusters in different proportions. Furthermore, the total contribution from each sample to final mixture of capsules should be equivalent. Therefore, we propose to model the procedure of dynamic routing as solving the following optimization question.
\begin{align}
\label{eq:opt}
\boldsymbol{v},\boldsymbol{r}=\mathop{\arg\max}_{\boldsymbol{v}, \boldsymbol{r}}\hat{f}(\boldsymbol{v},\boldsymbol{r}) \qquad s.t.\quad\forall{i,j}:r_{ij}>0,\sum\limits_{j=1}^{n_{l+1}}r_{ij}=1.
\end{align}

We will propose two different strategies to solve eq.~\ref{eq:opt} in the following parts, and discuss the relationship between the proposed methods and other existing routing methods.

\subsection{Routing based on mean shift}
Mean shift~\cite{comaniciu2002mean} is a typical clustering method based on the framework of KDE. It is extensively applied in the fields of feature analysis and related vision tasks such as segmentation and object tracking~\cite{meanshift}. The original mean shift method iteratively maximizes $\hat{f}(\boldsymbol{v})$ in eq.~\ref{eq:kde} by solving the following equation
\begin{align}
\label{eq:giter}
\triangledown\hat{f}(\boldsymbol{v})=\frac{1}{nz_k}\sum\limits_{i=1}^nk'(d(\boldsymbol{v}-\boldsymbol{u}_i))\frac{\partial{d}(\boldsymbol{v}-\boldsymbol{u}_i)}{\partial{\boldsymbol{v}}}=0.
\end{align}

By replacing $\hat{f}(\boldsymbol{v})$ with $\hat{f}(\boldsymbol{v},\boldsymbol{r})$, we propose to optimize variables $\boldsymbol{v}$ and $\boldsymbol{r}$ alternately. First, $\boldsymbol{v}_j^{\tau}$ can be updated by analogously solving eq.~\ref{eq:giter} with fixed $r_{ij}^{\tau}$ as
\begin{align}
\label{eq:iterv}
\boldsymbol{v}_j^{\tau+1} = \frac{\sum\limits_{i=1}^{n_l}r_{ij}^{\tau}a_i^uk'(d(\boldsymbol{v}_j^{\tau}-\boldsymbol{u}_i))\boldsymbol{u}_i}{\sum\limits_{i=1}^{n_l}r_{ij}^{\tau}a_i^uk'(d(\boldsymbol{v}_j^{\tau}-\boldsymbol{u}_i))}.
\end{align}

This form of $\boldsymbol{v}_j$ intuitively shows that the cluster can be explained as normalized weighted summation of candidate samples. The weight consists of the coefficient $r_{ij}$, the prior knowledge of candidate sample and the derivate of kernel function. 

Then, with updated $\boldsymbol{v_j^{\tau}}$, we optimize $r_{ij}$ with the standard gradient descent method as
\begin{align}
\label{eq:iterr}
r_{ij}^{\tau+1}=r_{ij}^{\tau}+\alpha{a}_i^uk(d(\boldsymbol{v}_j^{\tau}-\boldsymbol{u}_i)),
\end{align}
where $\alpha$ is the hyper parameter to control step size. In experiments we directly use $\alpha=1$ as a common configuration. To satisfy the constraints in eq.~\ref{eq:opt}, we simply normalize $r_{ij}$ as $r_{ij}=r_{ij}/\sum_{j=1}^{n_{l+1}}r_{ij}$ for further calculation. Combining eq.~\ref{eq:iterv} and eq.~\ref{eq:iterr}, a dynamic routing method can be obtained as algorithm~\ref{alg:grouting}.
\begin{algorithm}
  \caption{Dynamic routing based on mean shift.}
  \label{alg:grouting}
  \begin{algorithmic}
   \REQUIRE poses $\boldsymbol{u}_{i}$, activations $a_i^u$
   \STATE Initialize $\forall{i,j}$: $r_{ij}=1/n_{l+1}$
   \FOR{$r$ iterations}
   \STATE 1. $\forall{i,j}$: $r'_{ij}\gets{\frac{r_{ij}}{\sum_jr_{ij}}}$
   \STATE 2. $\forall{j}$: $\boldsymbol{v}_j\gets{\frac{\sum_ir'_{ij}a_i^uk'(d(\boldsymbol{v}_j-\boldsymbol{u}_{i}))\boldsymbol{u}_{i}}{\sum_ir'_{ij}a_i^uk'(d(\boldsymbol{v}_j-\boldsymbol{u}_{i}))}}$
   \STATE 3. $\forall{i,j}$: $r_{ij}\gets{r}_{ij}+a_i^uk(d(\boldsymbol{v}_j-\boldsymbol{u}_{i}))$

   \ENDFOR
   \RETURN capsules with poses $\boldsymbol{v}_j$
  \end{algorithmic}
\end{algorithm}

Here we omit $\alpha$ in algorithm~\ref{alg:grouting}. Note that the proposed routing method maximizes $\hat{f}(\boldsymbol{v},\boldsymbol{r})$ by following a framework of coordinate descent~\cite{wright2015coordinate}. Variables $\boldsymbol{v}$ and $\boldsymbol{r}$ are optimized alternately towards the direction of partial gradient. The value of $\hat{f}(\boldsymbol{v},\boldsymbol{r})$ is ensured to get increased or unchanged after each iteration. Exploiting different kernel functions and distance metrics in algorithm~\ref{alg:grouting} can lead to specific instances of routing methods.

\subsection{Routing based on expectation maximization}
Eq.~\ref{eq:opt} can be optimized within an Expectation Maximization (EM) framework~\cite{dempster1977maximum} as well. Due to the symmetry of kernel function, $k(d(\boldsymbol{v}_j-\boldsymbol{u}_i))$ can also be explained as the likelihood of sample $\boldsymbol{u}_i$ given the assumption of variable $\boldsymbol{v}_j$. From this point of view, Eq.~\ref{eq:kdem} can be treated as an approximation of the log-likelihood function given samples from the mixture model which consists of $\boldsymbol{v}_j$ as components and $r_{ij}$ as hidden weights. The EM algorithm can be exploited here to maximize $\hat{f}(\boldsymbol{v},\boldsymbol{r})$ by alternately optimizing $\boldsymbol{v}_j$ and updating $r_{ij}$ as its expectation. In analog to the standard EM algorithm, we explicitly introduce the mixture coefficient $\pi_j$ to calculate the expectation of $r_{ij}$, getting algorithm~\ref{alg:emrouting} as another strategy to solve eq.~\ref{eq:opt}.

\begin{algorithm}
  \caption{Dynamic routing based on EM algorithm.}
  \label{alg:emrouting}
  \begin{algorithmic}
   \REQUIRE poses $\boldsymbol{u}_{i}$, activations $a_i^u$
   \STATE Initialize $\forall{i,j}$: $r_{ij}=1/n_{l+1}$
   \FOR{$r$ iterations}
   \STATE 1. $\forall{i,j}$: $r'_{ij}\gets{\frac{r_{ij}}{\sum_jr_{ij}}}$
   \STATE 2. $\forall{j}$: $\boldsymbol{v}_j\gets{\frac{\sum_ir'_{ij}a_i^u\boldsymbol{u}_{i}}{\sum_ir'_{ij}a_i^u}}$
   \STATE 3. $\forall{j}$: $\pi_j\gets{\frac{\sum_ir'_{ij}}{\sum_j\sum_ir'_{ij}}}$
   \STATE 4. $\forall{i,j}$: $r_{ij}\gets{\pi_j}k(d(\boldsymbol{v}_j-\boldsymbol{u}_{i}))$
   \ENDFOR
   \RETURN capsules with poses $\boldsymbol{v}_j$
  \end{algorithmic}
\end{algorithm}

Algorithm ~\ref{alg:emrouting} basically follows the standard EM algorithm to maximize $\hat{f}(\boldsymbol{v},\boldsymbol{r})$. Comparing with another well-known application scenario of EM algorithm, the Gaussian Mixture Model (GMM), our proposed mixture model based on KDE can be treated as a simplified version of standard GMM. The simplification mainly comes from that KDE is based on nonparametric kernel function, rather than the normal distribution configured by expectation $\boldsymbol{\mu}$ and variance $\boldsymbol{\Sigma}$. Hence the calculation of $k(d(\boldsymbol{v}_j-\boldsymbol{u}_{i}))$ requires much less computation than Gaussian function $N(\boldsymbol{\mu},\boldsymbol{\Sigma})$. Also, the update of model in step 2 of algorithm~\ref{alg:emrouting} only requires calculation about $\boldsymbol{v}$ without the variance $\boldsymbol{\Sigma}$ as in GMM. We will empirically prove that such simplification can still provide acceptable performance for dynamic routing.

\textbf{Activation of capsule.} For generic machine learning tasks, the activation of capsule is required as final result to reflect the magnitude of current capsule. It also appears in the routing procedure at above layer as prior knowledge. For the resulting capsule at layer $l+1$, we propose a unified form of activation $a_j^v$ for both routing methods as
\begin{align}
\label{eq:ajv}
a_j^v\triangleq{softmax}(\sum\limits_{i=1}^{n_l}r'_{ij}a_i^u(k(\sum\limits_{d=1}^Dd(u_{id}-\beta_{jd}v_{jd})+\beta_{j0})),
\end{align}
where $r'_{ij}$ is the normalized version of $r_{ij}$ as the result of step 1 in algorithm~\ref{alg:grouting} and~\ref{alg:emrouting}. Here the distance metric is calculated at each dimension $d$ separately. If we ignore the linear coefficients $\boldsymbol\beta_j\in\mathbb{R}^{D+1}$, one can see from eq.~\ref{eq:kdem} that activation is the absolute value of resulting density at $\boldsymbol{v}_j$ after routing. It is consistent with the original purpose of combining activation to capsule. Parameters $\boldsymbol\beta\in\mathbb{R}^{D+1}$ at each dimension are learned by standard back propagation to provide a linear combination rather than the rigid connection between pose and activation. Finally, a softmax function is exploited here as the guarantee of a probability distribution over activations.

\textbf{Relationship between two routing methods.} We propose two dynamic routing methods by maximizing the weighted KDE function $\hat{f}(\boldsymbol{v}, \boldsymbol{r})$ with different strategies. The proposed methods can be instantiated with different kernel functions and distance metrics. For generic cases without any specific background knowledge, we propose to directly adopt Epanechnikov kernel for simplification. Note that the Epanechnikov kernel can result in an identical format of step 2 in both algorithm~\ref{alg:grouting} and~\ref{alg:emrouting}. From this point of view, the only difference between two routing methods is the strategy for update of weights $\boldsymbol{r}$. One is based on gradient descent while the other is expectation.

\section{Network architecture}
\begin{figure}
\centering
\includegraphics[width=12cm]{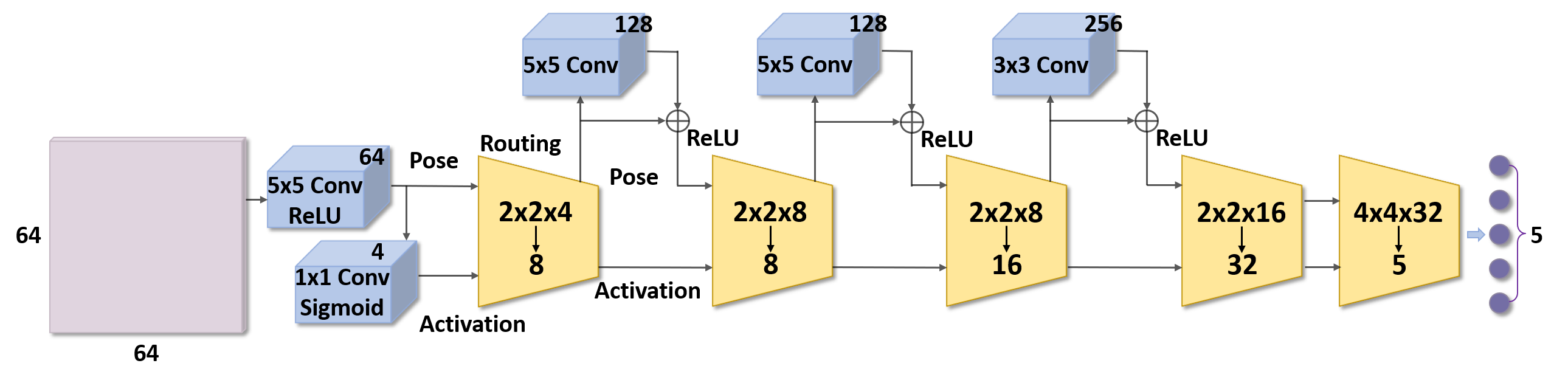}
\caption{The proposed network architecture for a $64\times{64}$ image. It contains 5 capsule layers where dynamic routing occurs. Except the last one, the dynamic routing at each capsule layer takes place within a $2\times{2}$ field with a stride of 2. At each block for dynamic routing, the number of input capsules is listed as width$\times$height$\times$number of capsules at each position. The number of output capsules is listed under the arrow. The side lengths of feature maps after each capsule layer are 32, 16, 8 and 4, respectively. The final dynamic routing takes all capsules within a $4\times{4}$ feature maps into account, resulting in 5 capsules as output. Here we take the smallNORB dataset for instance.}
\label{fig:network}
\end{figure}

Dynamic routing has already been proven as an effective mechanism for classification tasks on datasets with small images such as MNIST~\cite{lecun1998gradient} and smallNORB~\cite{lecun2004learning}. Our main concern here is about the efficiency and generality of dynamic routing. According to the observation of our implementation, they are the main bottleneck preventing dynamic routing from widespread utilization in practical vision tasks. Hence, we designed a relatively big network architecture for $64\times{64}$ image as in Figure~\ref{fig:network}. 

The proposed network is a hybrid architecture of convolutional layers and capsules with dynamic routing. For all convolutional layers, we adopt a stride of 1 for sliding kernels, and a padding of features to keep the feature map size. The network starts from convolutional layers for initialization of poses and activations of capsules. Then the primary dynamic routing takes place to represent 16 capsules with 8 capsules at every $2\times{2}$ field with a stride of 2. The feature maps are downsampled here with doubled number of capsules at every position.

For the rest part of the network, we stacked a homogeneous structure sequentially as shown in Figure~\ref{fig:network}. The proposed structure consists of a capsule layer and a residual block. The residual block is analog to the structure in ResNet~\cite{he2016deep}. It takes poses from below layer as input. The poses are summed with their residual maps to ReLU functions as output. We exploit such structure to integrate convolutional layer and capsules as a generic building block for deeper and larger networks. The feature map will only be downsampled at capsule layer, along with the increase of capsules at each position. The residual structure ensures an identical or better version of pose to adapt the propagation of corresponding activation in the following dynamic routing. We have tested different structures in our experiments, founding that the proposed residual block ensures most stable training for stack of capsules, especially in deeper networks.

We compose the pose of all capsules with $4\times{4}$ matrix as~\cite{hinton2018emrouting}. It requires many fewer parameters for transformation matrices during routing. This structure of pose leads to a network as in Figure~\ref{fig:network} with 1.2M parameters, in which nearly 90K parameters come from dynamic routing procedure. Differing from the stack of ``ConvCaps'' in~\cite{hinton2018emrouting}, our network shares parameters for routing at different positions without overlap between neighboring receptive fields. So although our proposed network has more parameters than network in~\cite{hinton2018emrouting}, it consumes less time for both training and inference.

For comparison, we also implement an analogous CNN to the architecture in Figure~\ref{fig:network} as baseline. The network consists of 5 convolutional layers without residual structure and a global average pooling to 5-way softmax output. Except for the $1\times{1}$ filters for initialization of activations, the structures of other 4 convolutional layers in Figure~\ref{fig:network} are all shared into the baseline CNN. The 5th convolutional layer contains a $3\times{3}$ kernel with 5 channels and a stride of 1. Capsule layers are replaced by direct max pooling between convolutional layers. All hidden layers are connected to the ReLU non-linearity. We designed the baseline CNN with an approximately equivalent amount of parameters as well as similar structures to our proposed network, ensuring a fair comparison between them.

\section{Experiments}
\subsection{Implementation details}
We compared three routing methods and a CNN baseline in different benchmarks. Here the Fast Routing based on Mean Shift (FRMS), Fast Routing based on EM (FREM), and EM routing from~\cite{hinton2018emrouting} are implemented within the proposed network architecture for consideration. During the training of these routing methods, we also integrate the reconstruction error from pose of resulting capsule as regularization in loss function. The structure of decoder from pose to original image is analogous to~\cite{sabour2017dynamic}. We resize the input to $32\times{32}$ image as ground truth in reconstruction to control the scale of fully connected layers.

In practical implementation, we use softmax function instead of standard normalization for the calculation of $r'_{ij}$ in step 1 of above algorithms. This modification can relax the kernel bandwidth restriction in KDE with only trivial increase of computation. With these modifications, we adopt the Epanechnikov kernel and $\ell_1$ norm: $d(\boldsymbol{v}-\boldsymbol{u}_i)=|\boldsymbol{v}-\boldsymbol{u}_i|$ as the default configuration for all experiments.

\subsection{Evaluation on smallNORB}
We started our experiments from comparison of different routing methods on smallNORB dataset, which contains 5 categories of targets with an amount of 24,300 $96\times{96}$ images. During training we randomly cropped $64\times{64}$ patches and augment data with random brightness and contrast. During test we directly cropped a $64\times{64}$ patch from the center of image. The image is centralized with unit deviation as input to the network. We implemented all the algorithms with TensorFlow 1.2.0~\cite{abadi2016tensorflow} on a hardware with Intel Xeon E5-2680 CPU at 2.4GHz and NVIDIA Tesla P40 GPU. We train the network for 50 epochs with mini-batches of size 50. 

We report the results of different methods in Table~\ref{tab:norb}. All the models are trained with 2 routing iterations for clustering. The best result is achieved by exploiting FREM within our proposed network architecture. The test error 2.2\% is in-par with state-of-the-art performance gained by another routing method~\cite{hinton2018emrouting}. Note that our method achieves the results at a higher resolution with much higher efficiency. 

From the perspective of efficiency, one can see that although routing methods consume more time for both training and inference than baseline CNN, they significantly reduce the test error by 80\% with similar network architecture. Moreover, the FREM and FRMS methods reduce the time consumption by nearly 40\% from EM routing. FREM slightly outperforms EM routing with our architecture.
\begin{table}
\centering
\caption{Performance evaluation on smallNORB.}
\label{tab:norb}
\begin{tabular}{cccc}
\toprule
Method & Inference time$\quad$ & Training time$\quad$ & Test error rate  \\
\midrule
FRMS & $0.158\pm{0.001}$s$\quad$ & $0.470\pm{0.002}$s$\quad$ & 2.6\% \\
FREM & $0.158\pm{0.003}$s$\quad$ & $0.471\pm{0.002}$s$\quad$ & 2.2\%\\
EM routing & $0.252\pm{0.003}$s$\quad$ & $0.744\pm{0.003}$s$\quad$ & 2.3\%\\
Baseline CNN & $0.043\pm{0.003}$s$\quad$ & $0.064\pm{0.001}$s$\quad$ & 11.3\%\\
\bottomrule
\end{tabular}
\end{table}

\textbf{Ablation study.} We evaluated the influences of varying different configurations to routing methods on smallNORB in Table~\ref{tab:ablation}. One can see that different number of iterations for dynamic routing severely impacts the final performance of models. Setting iteration number to 1 leads to failures of training for all methods. For the number of iterations as 3, we observed a side effect brought by long paths between capsules. We assume that this is because the training of transformation matrices relies on the back propagation of gradients from every $\boldsymbol{u}_i$ at different step of iterations. Routing with more iterations tends to be impacted by the vanishing gradient problem~\cite{he2016deep}. Hence, we recommend to use 2 iterations as the default configuration for routing.

Meanwhile, we also evaluated the influence of different initialization configurations. We adopt the Truncated Normal Distribution (TND) with 0 mean and 0.01 standard deviation as the common initialization for all trainable weights. However, for transformation matrices, we tried higher standard deviation as initialization. One can see from Table~\ref{tab:ablation} that, TND with standard deviation as 0.1 or 1.0 provides much better performances than 0.01. We assume that higher deviation can lead to more differences between transformation matrices, which can serve as better initialization of the clustering.
\begin{table}
\centering
\caption{Ablation study on smallNORB. Here we omit variances of time consumption for simplicity. ``Stddev'' represents the standard deviation of TND for initialization.}
\label{tab:ablation}
\begin{tabular}{cccc}
\toprule
Method & Routing iterations & Stddev & Test error rate \\
\midrule
FRMS & 1 & 0.1 & 76.9\% \\
FRMS & 2 & 0.01 & 5.9\% \\
FRMS & 2 & 0.1 & 2.6\% \\
FRMS & 2 & 1.0 & 2.7\% \\
FRMS & 3 & 0.1 & 7.1\% \\
FREM & 1 & 0.1 & 77.8\% \\
FREM & 2 & 0.01 & 5.6\% \\
FREM & 2 & 0.1 & 2.2\% \\
FREM & 2 & 1.0 & 2.3\% \\
FREM & 3 & 0.1 & 6.0\% \\
EM routing & 1 & 0.1 & 70.2\% \\
EM routing & 2 & 0.01 & 6.2\% \\
EM routing & 2 & 0.1 & 2.3\% \\
EM routing & 2 & 1.0 & 2.3\% \\
EM routing & 3 & 0.1 & 5.8\% \\
\bottomrule
\end{tabular}
\end{table}

\begin{figure}
\centering
\includegraphics[width=12cm]{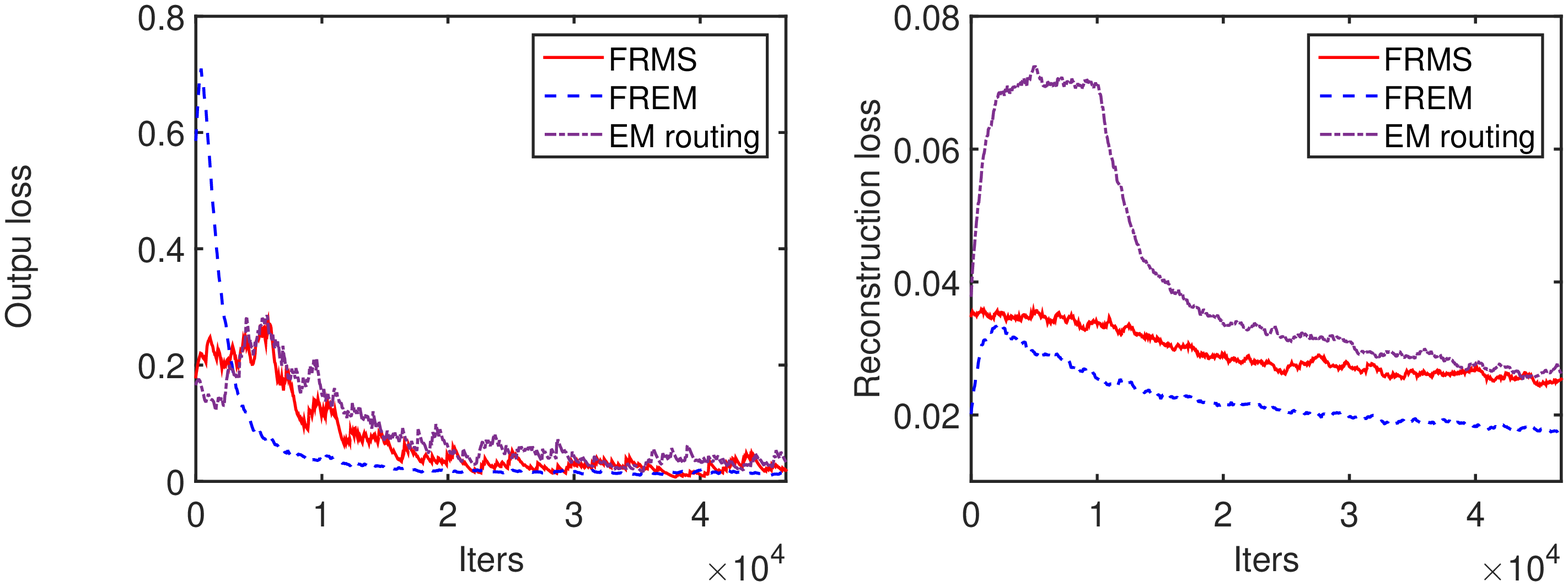}
\caption{The tendencies of output loss and reconstruction loss.}
\label{fig:loss}
\end{figure}

\textbf{Training loss.} We illustrate the curves of training loss in Figure~\ref{fig:loss}. Our training loss consists of output loss defined on activations and reconstruction loss defined on poses. For the output loss function, we tried margin loss~\cite{sabour2017dynamic} and spread loss~\cite{hinton2018emrouting} for all candidate methods. We implement the spread loss by increasing the margin from 0.2 to 0.9 within 5 epochs. It turns out FREM can be well trained with both kinds of loss functions, while the training of FRMS and EM routing seriously relies on the relaxation from spread loss to ensure the diversity of capsules.

\subsection{Other datasets}
We also tested the proposed methods on MNIST, Fashion-MNIST~\cite{xiao2017online} and CIFAR10~\cite{krizhevsky2009learning}. To adapt the size of images in these datasets, we removed the second capsule layer as well as the residual block from the network in Figure~\ref{fig:network}. With four capsule layers the network can process an input image with $32\times{32}$ pixels. We resize all images to this resolution in the experiments.

We listed all results on different datasets for comparison in Table~\ref{tab:datasets}. For MNIST, all methods approximately achieve the same accuracy, since the results are nearly saturated. For Fashion-MNIST, our methods outperform another implementation of routing-by-agreement method with 8M parameters~\cite{xiao2017online}. The FRMS method slightly outperforms the FREM on this dataset. For CIFAR10, we modified the first $5\times{5}$ convolutional layer of network as 256 output channels with 3 color channels as input. The reported results are achieved by an ensemble of 5 models with the same hyper parameters as in the experiments on smallNORB. We omit the comparison of time consumptions of methods here since it is basically consistent to Table~\ref{tab:norb}. In the case of lower resolution, there is a trivial gap between our methods and the EM routing. However, our efficient methods for routing still show high potentiality to prompt the performance of baseline CNN.   
\begin{table}
\centering
\caption{Results of proposed methods on datasets.}
\label{tab:datasets}
\begin{tabular}{cccc}
\toprule
Method & MNIST & Fashion-MNIST & CIFAR10 \\
\midrule
FRMS & 0.42\% & 6.0\% & 15.6\% \\
FREM & 0.38\% & 6.2\% & 14.3\% \\
EM routing & 0.32\% & 5.8\% & 11.6\% \\
Baseline CNN & 0.65\% & 7.6\% & 19.2\% \\
\bottomrule
\end{tabular}
\end{table}

\begin{figure}
\centering
\includegraphics[width=12cm]{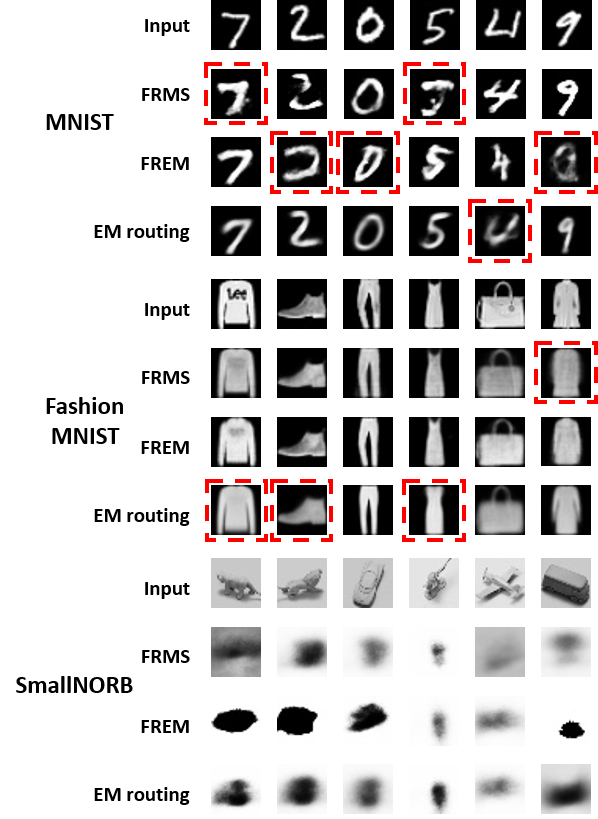}
\caption{Samples of reconstructions with different routing methods on datasets. For MNIST and Fashion-MNIST, we subjectively signed some reconstruction results with red dashed rectangle as reference. Some of these samples lack of details while others show confusing features for classification.}
\label{fig:recon}
\end{figure}

We also illustrate some reconstruction samples on different datasets in Figure~\ref{fig:recon}. One can see that routing methods result in reasonable reconstruction on MNIST and Fashion-MNIST. In contrast, they can only reconstruct an implicit prior of input on smallNORB, despite that the reconstruction errors do reduce in Figure~\ref{fig:loss}. The training of model on smallNORB seems not impacted by this failure. We assume that smallNORB mainly consists of target models from different azimuths without any noise from background. The mechanism of routing can handle such affine transformation fairly well. We met the same failures of reconstruction on CIFAR10 as well. It seriously impacts the final performance of capsule networks. Structures with higher complexity rather than only fully connected layers in decoder could be a potential way to mitigate this problem.

\section{Discussion and Related Work}

We treat dynamic routing as an advanced compression of knowledge from capsules at layer $l$ to $l+1$. Features from candidate capsules are represented by a mixture of less capsules with corresponding magnitudes. The compression mainly relies on the unsupervised clustering as well as transformation matrices. Conventional clustering methods for mixture models normally start from randomized samples as initialization, while dynamic routing relies on transformation matrices as guarantee of diversity over different clusters. The learning of transformation matrices by back propagation thus can be modeled as the procedure of compressing knowledge into new clusters for a more efficient representation. Based on these principles, we proposed two simplified routing methods for better cooperation with back propagation learning. Our methods can also be treated as a generalized version of existing routing methods.

\textbf{FRMS and routing-by-agreement~\cite{sabour2017dynamic}.} We use the term ``Routing-by-agreement'' to represent the original routing method in~\cite{sabour2017dynamic}. By a non-linear ``squashing'' function, all lengths of capsules here are compressed into $[0, 1)$ in analog to probabilities. Since routing-by-agreement is only deployed once at the ``DigitCaps'' layer, no prior from below layer is considered, i.e. $a_i^u$ can be omitted from algorithm~\ref{alg:grouting}. With the same modification, it can be shown that by using the Epanechnikov kernel function and a non-strict distance metric from the cosine similarity as
\begin{align}
\label{eq:innerprod}
\langle{\boldsymbol{u}, \boldsymbol{v}}\rangle\triangleq{1}-\boldsymbol{u}^T\boldsymbol{v},
\end{align}
a quite similar routing method to routing-by-agreement can be derived from FRMS as algorithm~\ref{alg:variant}. Here we replace $d(\boldsymbol{u}-\boldsymbol{v})$ by $\langle{\boldsymbol{u},\boldsymbol{v}}\rangle$ due to its specific form. 
\begin{algorithm}
  \caption{Variant of FRMS.}
  \label{alg:variant}
  \begin{algorithmic}
   \REQUIRE poses $\boldsymbol{u}_{i}$, activations $a_i^u$
   \STATE Initialize $\forall{i,j}$: $r_{ij}=1/n_{l+1}$
   \FOR{$r$ iterations}
   \STATE 1. $\forall{i,j}$: $r'_{ij}\gets{softmax(r_{ij})}$
   \STATE 2. $\forall{j}$: $\boldsymbol{v}_j\gets{\sum_ir'_{ij}\boldsymbol{u}_i}$
   \STATE 3. $\forall{i,j}$: $r_{ij}\gets{r}_{ij}+\boldsymbol{u}_i^T\boldsymbol{v}_j$

   \ENDFOR
   \RETURN capsules with poses $\boldsymbol{v}_j$
  \end{algorithmic}
\end{algorithm}

Routing-by-agreement directly assigns length of pose as the activation of capsule without any linear coefficient $\boldsymbol\beta$. We can analogously modify the definition of activation in eq.~\ref{eq:ajv} as
\begin{align}
\label{eq:arba}
a_j^v\triangleq{softmax}(\sum\limits_{i=1}^{n_l}r'_{ij}\boldsymbol{u}_{i}^T\boldsymbol{v}_j).
\end{align}
Here we also remove $a_i^u$ from the equation for an easier comparison. Note that since we update $\boldsymbol{v}_j$ here as $\boldsymbol{v}_j\gets\sum_{i=1}^{n_l}r'_{ij}\boldsymbol{u}_i$, eq.~\ref{eq:ajv} can be rewritten as the squared length of $\boldsymbol{v}_j$ directly.

However, from eq.~\ref{eq:giter} we can see that the derivation of step 2 in algorithm~\ref{alg:variant} is suspicious due to the specific form of $\langle{\boldsymbol{u},\boldsymbol{v}}\rangle$. The convergence of whole algorithm is not well ensured. During our experiments, we also observed that stack of algorithm~\ref{alg:variant} can easily lead to failures of training. So we remove the method from comparison.

\textbf{FREM and EM routing~\cite{hinton2018emrouting}.} The proposed FREM can be treated as a simplified version of EM routing. The detailed analysis of the relationship between KDE based mean shift and EM could be referred to~\cite{4135673}. EM routing addresses the clustering question completely within the framework of GMM, maximizing the log-likelihood function with Gaussian function as the divergence metric. The differences between FREM and EM routing can be mainly summarized as follows: (1) EM routing exploits activation $a_j^v$ in the update of weight $r_{ij}$ as prior knowledge of cluster. We use the mixture coefficient $\pi_{j}$ as standard EM algorithm for simplification. According to our implementation, this modification only brings trivial impact to routing performance. (2) Gaussian function provides asymmetrical kernels with the variance $\boldsymbol{\Sigma}$ at each dimension of cluster. Our KDE based kernel is symmetrical at all dimensions. This simplification leads to non-trivial influence to the performance of our routing method. However, with the utilization of convolutional layers, the impact is significantly mitigated to an acceptable level. 



\section{Conclusion}
In this paper, we propose two efficient routing methods by generalizing existing methods within the framework of weighted KDE. Rather than constructing network with capsule structures independently for higher performance, we propose to exploit capsules and dynamic routing as effective complements with convolutional units. Experimental results show that such hybrid structure is promising to provide efficient solutions with much higher capacity for large scale vision tasks. In our future work, we plan to further prompt the performance of capsule networks on practical image datasets with higher resolution, e.g., STL-10~\cite{coates2011analysis}.
\bibliographystyle{splncs}
\bibliography{reference}
\end{document}